\documentclass[sigconf]{acmart} %%,anonymous

\usepackage{tabularx}
\usepackage{cleveref}
\usepackage{graphicx} 
\usepackage{subfigure}
\usepackage{multirow}

\AtBeginDocument{%
  }

\renewcommand\footnotetextcopyrightpermission[1]{}

\begin{document}
\pagestyle{empty} % 只保留页脚的页码，清除页眉
\newcommand{\ours}{DGAE}

\title{DGAE: Diffusion-Guided Autoencoder for Efficient Latent Representation Learning}
\author{Dongxu Liu}
\affiliation{%
    \institution{Institute of Automation, Chinese Academy of Sciences}
    \city{Beijing}
    \country{China}}

\author{Jiahui Zhu}
\affiliation{%
    \institution{Independent Researcher}
    \country{China}}

\author{Yuang Peng}
\affiliation{%
    \institution{Tsinghua University}
    \city{Beijing}
    \country{China}}

\author{Haomiao Tang}
\affiliation{%
    \institution{Tsinghua University}
    \city{Beijing}
    \country{China}}

\author{Yuwei Chen}
\affiliation{%
    \institution{Institute of Computing Technology, Chinese Academy of Sciences}
    \city{Beijing}
    \country{China}}

\author{Chunrui Han}
\affiliation{%
    \institution{StepFun}
    \city{Beijing}
    \country{China}}

\author{Zheng Ge}
\affiliation{%
    \institution{StepFun}
    \city{Beijing}
    \country{China}}

\author{Daxin Jiang}
\affiliation{%
    \institution{StepFun}
    \city{Beijing}
    \country{China}}

\author{Mingxue Liao}
\affiliation{%
    \institution{Institute of Automation, Chinese Academy of Sciences}
    \city{Beijing}
    \country{China}}

\authornote{Corresponding Authors.}

\begin{abstract}
Autoencoders empower state-of-the-art image and video generative models by compressing pixels into a latent space through visual tokenization. Although recent advances have alleviated the performance degradation of autoencoders under high compression ratios, addressing the training instability caused by GAN remains an open challenge.  While improving spatial compression, we also aim to minimize the latent space dimensionality, enabling more efficient and compact representations. To tackle these challenges, we focus on improving the decoder’s expressiveness. Concretely, we propose {\ours}, which employs a diffusion model to guide the decoder in recovering informative signals that are not fully decoded from the latent representation. With this design, {\ours} effectively mitigates the performance degradation under high spatial compression rates. At the same time, {\ours} achieves state-of-the-art performance with a $2\times$ smaller latent space. When integrated with Diffusion Models, {\ours} demonstrates competitive performance on image generation for ImageNet-1K and shows that this compact latent representation facilitates faster convergence of the diffusion model.
\end{abstract}


\maketitle
\section{Introduction}
\label{sec:intro}
\begin{figure}[htbp]
    \centering
    \subfigure[Scale up the discriminator.]{
        \includegraphics[width=0.35\textwidth]{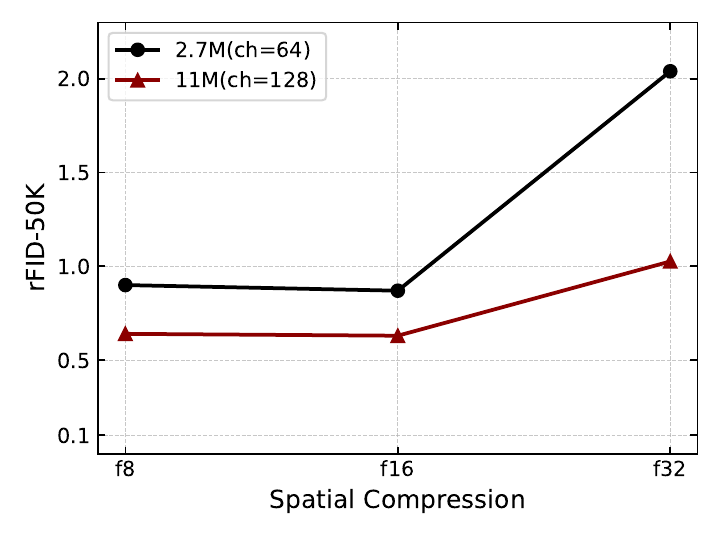}
        \label{fig:gan_scale}
    }
    \subfigure[Scale up the encoder and decoder.]{
        \includegraphics[width=0.35\textwidth]{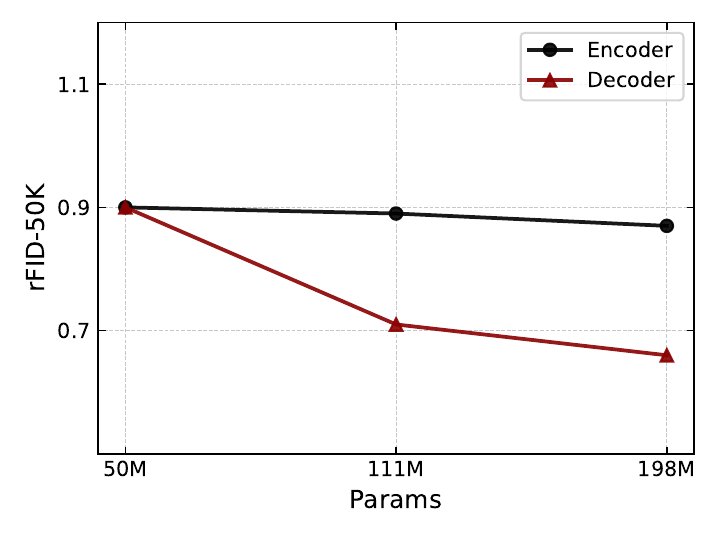}
        \label{fig:decoder_scale}
    }

    \caption{\textbf{(a)} Scaling up the discriminator in GANs can mitigate the decline in reconstruction accuracy of autoencoders under high spatial compression rates, while also enhancing reconstruction performance at low spatial compression rates. \textbf{(b)} Scaling up the decoder effectively improves the reconstruction quality of the autoencoder, while scaling up the encoder has little effect. }
    \label{fig:fid_sdvae_opt}
    \vspace{-15pt}
\end{figure}
Autoencoders serve as a foundational component in modern high-resolution visual generation. Their ability to compress vast, high-dimensional image data into a compact and information-rich latent space is crucial for the efficiency and success of subsequent generative processes, most notably demonstrated by Latent Diffusion Models (LDMs) ~\cite{rombach2022stable_diffusion, william2023dit, blattmann2023videoldm_align_your_latens, blattmann2023svd_stablevideodiffusion, bao2023uvit}. By operating within this lower-dimensional latent space, powerful models like diffusion can be trained and perform inference far more efficiently. The success of latent-based generative frameworks relies not only on powerful generative models but also significantly on the quality of the autoencoder, which constructs the latent space where generation occurs. A fundamental challenge lies in the inherent trade-off between spatial compression and reconstruction fidelity, as aggressive compression lowers computational cost but compromises the visual quality of autoencoder reconstructions ~\cite{rombach2022stable_diffusion}.

The trade-off between compression and reconstruction fidelity remains a non-trivial problem to resolve. DCAE~\cite{chen2024dcae} addresses this challenge by incorporating residual connections during both downsampling and upsampling stages, thereby achieving higher spatial compression rates without degrading reconstruction quality.
In this work, we investigate a complementary direction that focuses on the training objectives. As shown in \cref{fig:gan_scale}, we observe that the commonly used GAN loss~\cite{larsen2016autoencoding} not only enhances the perceptual quality of reconstructed images, but that increasing the discriminator capacity further mitigates the degradation typically caused by aggressive compression. We hypothesize that \textbf{a stronger discriminator provides richer learning signals, thus enhancing the expressiveness of the decoder.}

To validate this hypothesis, we independently scaled the parameter sizes of both the encoder and decoder. Interestingly, we found that \textbf{increasing decoder capacity yields substantial improvements in reconstruction quality, while enlarging the encoder has minimal impact}, as shown in \cref{fig:decoder_scale}. This result suggests that the decoder plays a dominant role in maintaining visual fidelity under high compression, indicating that future optimization efforts should prioritize decoder design.

Although GAN-guided VAEs have addressed the challenge of high spatial compression, they suffer from issues such as \textbf{mode collapse} and \textbf{sensitivity to hyperparameters}, making them less ideal for guiding the decoder to learn robust latent representations. In recent years, diffusion models have emerged as a dominant paradigm in visual generation due to their stable training dynamics and theoretically grounded framework. However, their potential in representation learning remains underexplored.

In this work, we propose {\ours}, a novel and stable autoencoder architecture that leverages a diffusion model~\cite{aapo2015scorematching, ho2020ddpm} to guide the decoder in learning a denser and more expressive latent space. As illustrated in \cref{fig:model_arch}, unlike GAN-guided methods such as SD-VAE~\cite{rombach2022stable_diffusion}, our approach conditions on the encoder’s latent representation and reconstructs the image by progressively denoising from random noise. The core idea is to \textbf{transfer the strong data modeling capabilities of diffusion models into the decoder of autoencoder}, thereby enhancing its ability to reconstruct high-frequency visual signals such as text and textures.

We demonstrate that {\ours} addresses the  reconstruction quality degradation and training instability observed in GAN-guided VAEs under high spatial compression, while also accelerating diffusion model training. Notably, {\ours} achieves comparable reconstruction performance to SD-VAE with a significantly smaller latent size. Furthermore, we show that optimizing for high spatial compression is not the sole objective in autoencoder design—\textbf{smaller latent representations can also facilitate faster convergence in downstream generative models (\cref{sec:latent_vis}).}

In summary, our contributions are as follows:
\begin{itemize}
    \item We analyze conventional autoencoder designs and empirically show that the decoder plays a critical role in determining reconstruction quality.
    \item We introduce {\ours}, a diffusion-guided autoencoder that achieves more compact and expressive latent representations.
    \item We demonstrate that smaller latent representations not only enable high compression but also accelerate training in diffusion-based generative models.
\end{itemize}

\section{Preliminaries}
\label{sec:background}
To facilitate comprehension of our work, we provide a concise overview of the continuous visual tokenization and diffusion model.
\subsection{VAEs}

Variational Autoencoders (VAEs)~\cite{kingma2014vae} introduce a \textbf{probabilistic framework} for learning latent representations by \textbf{modeling the underlying data distribution}. Given an image $\mathbf{X} \in \mathbb{R}^{H \times W \times 3}$, the encoder $q_\phi$ maps it to a latent representation $z \in \mathbb{R}^{\frac{H}{f} \times \frac{W}{f} \times c}$ using a downsampling factor $f$, and the decoder $p_\theta$ reconstructs the image $\hat{\mathbf{X}}$ from $z$.

\textbf{The core objective of a VAE is to maximize the Evidence Lower Bound (ELBO)}, which consists of two terms: (1) a \textbf{likelihood term} that encourages the decoder to assign high probability to the observed data given the latent variable, and (2) a \textbf{KL divergence term} that regularizes the latent distribution to match a prior, typically a standard Gaussian. The ELBO is defined as:
\begin{equation}
L(\theta, \phi) = \mathbb{E}_{q_{\phi}(z|x)}[\log p_{\theta}(x|z)] - \text{KL}(q_{\phi}(z|x) \| p(z))
\label{eq:vae_elbo}
\end{equation}
where $q_{\phi}(z|x)$ is the variational posterior approximated by the encoder, and $p_{\theta}(x|z)$ is the generative likelihood modeled by the decoder. The first term, also referred to as the \textbf{reconstruction term}, depends on the assumed form of $p_{\theta}(x|z)$—for example, \textbf{under the common assumption that $p_{\theta}(x|z)$ is a Gaussian distribution with fixed unit variance}, this term becomes equivalent to a mean squared error ($\ell_2$) loss. Consequently, the overall VAE training objective comprises the reconstruction loss $\mathcal{L}_{\text{REC}}(X, \hat{X})$ and the KL divergence loss $\mathcal{L}_{\text{KL}}$.

To enhance the visual quality of reconstructions, recent VAE variants have incorporated additional supervision. One such term is the perceptual loss $\mathcal{L}_{\text{LPIPS}}$, which utilizes feature maps extracted from a pretrained VGG network~\cite{johnson2016perceptual} to \textbf{improve perceptual similarity}. Another is the adversarial loss $\mathcal{L}_{\text{GAN}}$, which \textbf{refines texture details through PatchGAN-style training~\cite{isola2017patchgan}}. The full loss function of the autoencoder can be written as:
\begin{equation}
    \mathcal{L}_{VAE} = \alpha\mathcal{L}_{\text{REC}} 
    + \beta\mathcal{L}_{\text{KL}} 
    + \eta\mathcal{L}_{\text{LPIPS}} 
    + \lambda\mathcal{L}_{\text{GAN}}
    \label{eq:sdvae_loss}
\end{equation}
where $\alpha$, $\beta$, $\eta$, $\lambda$ are weighting coefficients that balance the contribution of each term.

\subsection{Diffusion Models}
Diffusion models~\cite{ho2020ddpm, song2020sde} are a class of likelihood-based generative models that synthesize data by learning to reverse a progressive noising process. Similar to VAEs, diffusion models aim to maximize the data likelihood, but they do so by \textbf{modeling the data distribution through a parameterized denoising process} rather than explicit latent variational inference. 

From a score-based perspective, these models learn to approximate the score function $\nabla_x \log p_t(x)$—i.e., the gradient of the log-density of the data corrupted by noise at time $t$. The forward process incrementally perturbs a clean image $\mathbf{x}_0$ into a noisy version $\mathbf{x}_T$ through a Markov chain, often using Gaussian noise. The reverse process then reconstructs the data by learning a sequence of denoising steps that recover $\mathbf{x}_0$ from $\mathbf{x}_T$.

Formally, given a perturbation kernel $q(\mathbf{x}_t|\mathbf{x}_0)$ that defines the forward process, the score-based model is trained to match the true score $\nabla_{\mathbf{x}_t} \log q(\mathbf{x}_t|\mathbf{x}_0)$ using a parameterized neural network $s_\theta(\mathbf{x}_t, t)$. This training objective is typically framed as a denoising score matching loss~\cite{aapo2015scorematching,Siwei2012IGscorematching,song2019sbmodel,song2020sde,Pascal2011denoisingautoencoder}: \begin{equation} \mathcal{L}_{\text{DSM}} = \mathbb{E}_{\mathbf{x}_0, t, \mathbf{x}t}\left[\left|s_\theta(\mathbf{x}_t, t) - \nabla_{\mathbf{x}_t} \log q(\mathbf{x}_t|\mathbf{x}_0)\right|^2\right]. \end{equation}

After training, data samples can be generated by solving a stochastic differential equation (SDE) or its discretized form using the learned score function. Compared to GANs, diffusion models offer \textbf{more stable training} and \textbf{better likelihood estimates}, making them an attractive alternative for generative modeling and image reconstruction tasks.

\section{Approach}

\begin{figure}
  \centering
  \includegraphics[width=1.0\linewidth]{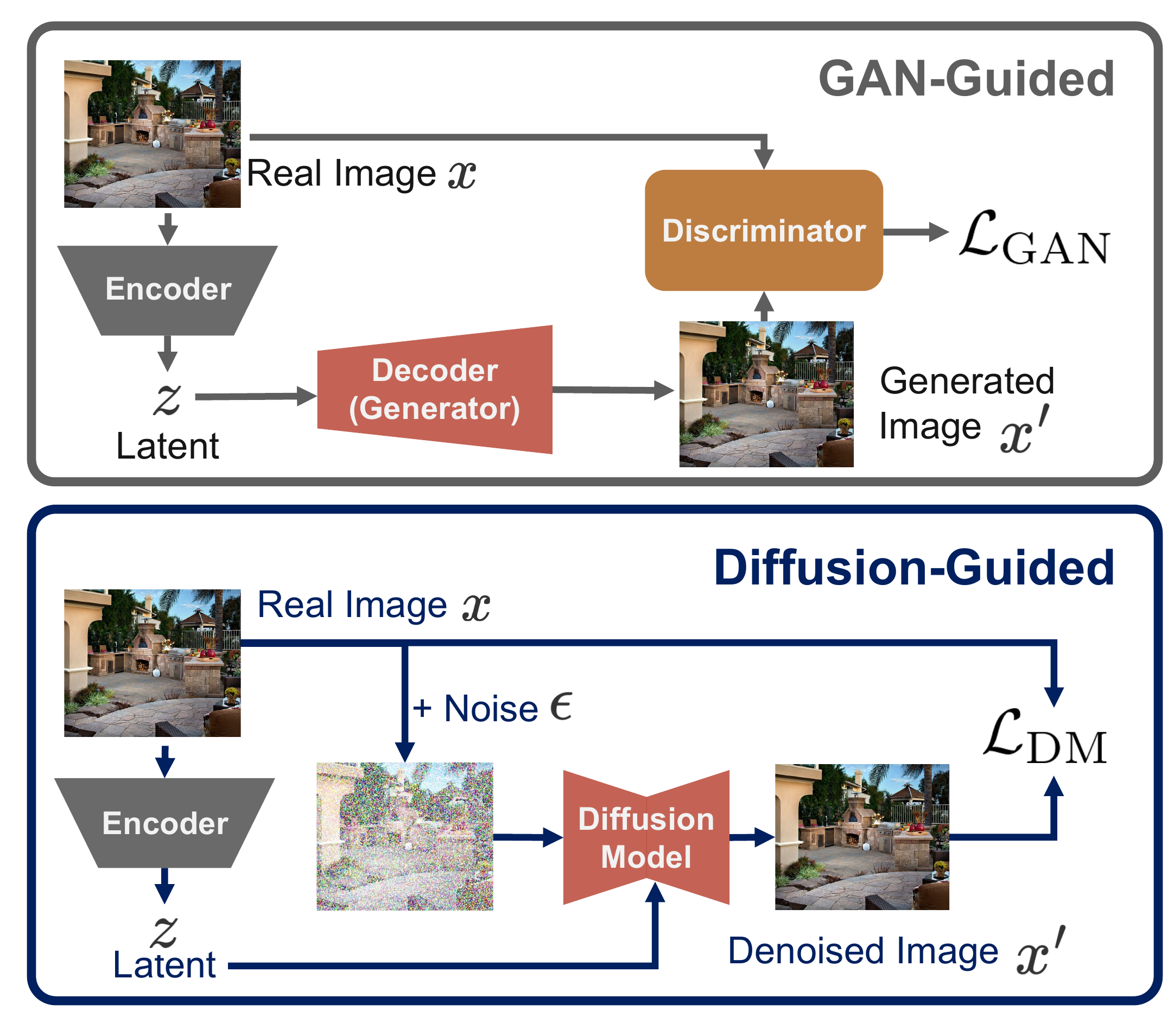}
  \caption{{\ours} is a diffusion-guided autoencoder, which is dedicated to enhancing the decoding capability of the decoder. Compared with GAN-guided methods, the latent representation \( z \) is no longer used for direct image reconstruction. Instead, it serves as a supervisory signal for the decoder, thereby better constraining \( p(x|z) \) to the data distribution \( p(x) \). }
  \label{fig:model_arch}
  \vspace{-15pt}
\end{figure}

Our primary innovation demonstrates that visual reconstruction is inherently a generative task, and the supervision of stronger generative models can significantly boost the decoder decoding capabilities. \Cref{fig:model_arch} illustrates the training process of {\ours}.

\subsection{From Gaussian to Diffusion Decoder}
In a standard VAE, the reconstruction term of the ELBO is defined as the expected log-likelihood of the observed data under the decoder distribution $p_\theta(x|z)$. The first term in ~\cref{eq:vae_elbo} measures how well the decoder can reconstruct the original input $x$ from the latent variable $z$ sampled from the encoder. In practice, this expectation is estimated by Monte Carlo sampling with $J$ samples, leading to
\begin{equation}
\mathbb{E}_{q_\phi(z|x)}[\log p_\theta(x|z)] \approx \frac{1}{J} \sum_{j=1}^{J} \log p_\theta(x | z_j),~~\text{where}~z_j \sim q_\phi(z|x).
\end{equation}

To make this objective tractable, conventional VAE methods typically assume the decoder distribution $p_\theta(x|z)$ to be an isotropic Gaussian with fixed variance, i.e.,
\begin{equation}
p_\theta(x \mid z)=\mathcal{N}\left(x ; \mu_\theta(z), \sigma^2 I\right) .
\end{equation}
Under this assumption, maximizing $\log p_\theta(x|z)$ becomes equivalent to minimizing an $\ell_2$ reconstruction loss between $x$ and $\hat{x} = \mu_\theta(z)$:
\begin{equation} \mathcal{L}_{\text{REC}} = |x - \hat{x}|_2^2. \end{equation}

However, this \textbf{Gaussian assumption imposes limitations on the expressiveness of the decoder}, especially for modeling complex, high-frequency structures such as textures and detailed semantics.

To overcome this limitation, we \textbf{replace the Gaussian decoder with a conditional diffusion model}, thereby removing the restrictive Gaussian assumption and allowing the model to directly learn the score function $\nabla_x \log p(x|z)$. Consequently, we reinterpret this expectation as being indirectly maximized via the following score-based surrogate objective:
\begin{equation}
\mathcal{L}_{\text{DSM}}=\mathbb{E}_{q\left(x_t \mid x\right)}\left[\lambda(t)\left\|s_\theta\left(x_t, t, z\right)-\nabla_{x_t} \log q\left(x_t \mid x\right)\right\|^2\right]
\end{equation}
which serves as a proxy for maximizing $\log p_\theta(x | z)$ without assuming a tractable likelihood form.

In this setup, \textbf{the decoder becomes a denoising network trained to reverse a fixed forward noising process conditioned on the latent variable $z$.} That is, given a clean image $x$ and its corresponding latent representation $z$, we perturb $x$ into a noisy sample $x_t$ using a diffusion process $q(x_t|x)$, and then train a conditional score network $s_\theta(x_t, t, z)$ to predict the gradient of the log-likelihood at each noise level as shown in \cref{fig:model_arch}.

\subsection{Training Objectives}
The above formulation replaces the original $\ell_2$ reconstruction loss with a denoising score matching loss, which can be interpreted as \textbf{a likelihood-based training objective under the score-based generative modeling framework}. Consequently, our model benefits from a more expressive and theoretically grounded decoder, significantly improving its ability to reconstruct fine-grained structures while maintaining the probabilistic rigor of the ELBO formulation.

In addition to the score-based loss, we find that incorporating the \textbf{perceptual loss further enhances the perceptual quality of reconstructed images.} However, since our reconstruction is now guided by a diffusion process, we must adapt the perceptual loss to align with our score-based training procedure. Specifically, during training, the model performs single-step predictions of the clean image via $x_0' = x_t - t \cdot v_\theta(x_t,t,z)$, where $v_\theta$ is the predicted noise (or velocity). To make perceptual loss compatible with this formulation, we compute it between the predicted $x_0'$ and the ground-truth image $x$, effectively supervising the model with perceptual feedback at each timestep.

Therefore, our final training objective combines the denoising score matching loss and the perceptual loss, and is defined as follows:
\begin{equation}
    \mathcal{L}_{DGAE} = \alpha\mathcal{L}_{\text{DSM}} 
    + \beta\mathcal{L}_{\text{KL}} 
    + \eta\mathcal{L}_{\text{LPIPS}} 
    \label{eq:dgae_loss}
\end{equation}

\subsection{Architecture}

% \subsection{Diffusion-Guided Autoencoder}
\textbf{Encoder.} Similar to SD-VAE, {\ours} employs a convolutional network architecture to map the input image $x$ to the latent representation $z$. The distribution of $z$ is as follows:
\begin{equation}
    q_\phi(z|x) = \mathcal{N}(z; \mu_\phi(x), \sigma_\phi(x)^2)
\end{equation}
where $\sigma_\phi(x)$, $\mu_\phi(x)$ are obtained by splitting the encoder's output $f_{\phi}(x)$. Then, $z$ is sampled from $q_\phi(z|x)$ through reparameterization.

\begin{figure*}[t!]
  \centering
  \includegraphics[width=1.0\linewidth,height=0.7\textheight, keepaspectratio]{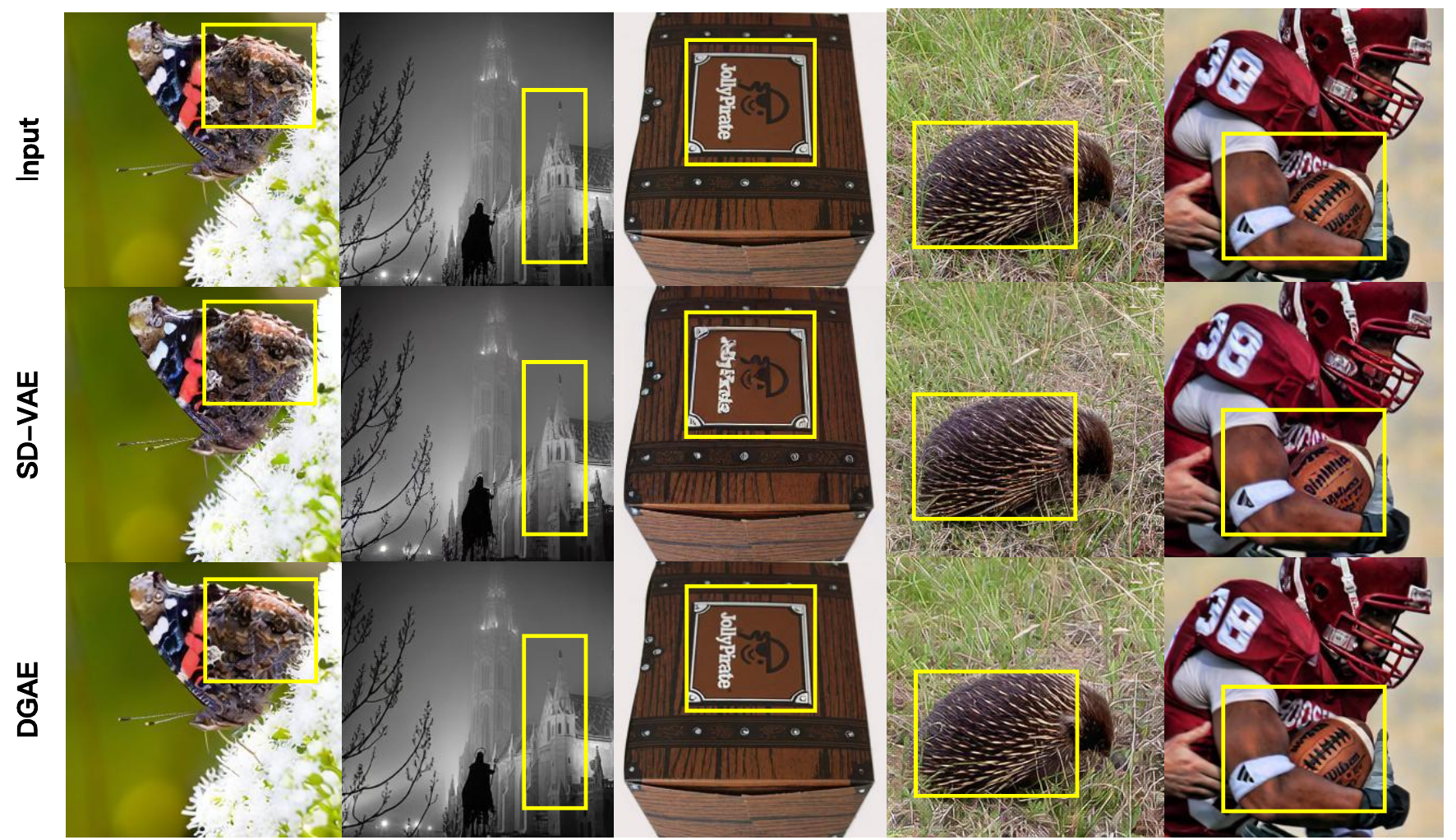}
  \caption{\textbf{Reconstructed samples of {\ours} and SD-VAE.} These results suggest that, despite employing a simpler combination of losses, {\ours} benefits from the strong modeling capacity of the diffusion decoder, leading to more effective recovery of fine-grained details such as textures and structural patterns.}
  \label{fig:sample_dgvae_vs_sdvae}
  \vspace{-10pt}
\end{figure*}

\noindent\textbf{Decoder.} Unlike previous deterministic decoding, the decoding task in {\ours} begins with a random noise. Specifically, the diffusion process utilizes latent representations  $z$ as conditional information, gradually denoising random noise $x_T$ to the original image $\hat{x}$:
\begin{equation}
    p_\theta(\hat{x}|z) = p(x_T)\prod_{t=1}^T p_\theta(\hat{x}_{t-1}|\hat{x}_{t}, z)
\end{equation}
where $\hat{x}_t$ represents the reconstructed image at time step $t$. The latent representation $z$ constrains the generation results of the diffusion process to the data distribution of the input image, while the iterative denoising of the diffusion process enhances the autoencoder's ability to model high-frequency details and local structures. 

\section{Experiments}
\label{sec:experiments}
To validate the effectiveness of {\ours}, we begin by outlining the experimental setup (\cref{sec:setup}). We then assess the reconstruction performance of {\ours} (\cref{sec:rec}) and examine the effectiveness of its learned latent space for diffusion models (\cref{sec:gen}). Finally, we analyze why {\ours} outperforms SD-VAE in \cref{sec:latent_vis}.
%-------------------------------------------------------------------------\
\begin{figure*}
  \centering
  \includegraphics[width=1.0\linewidth,height=0.7\textheight, keepaspectratio]{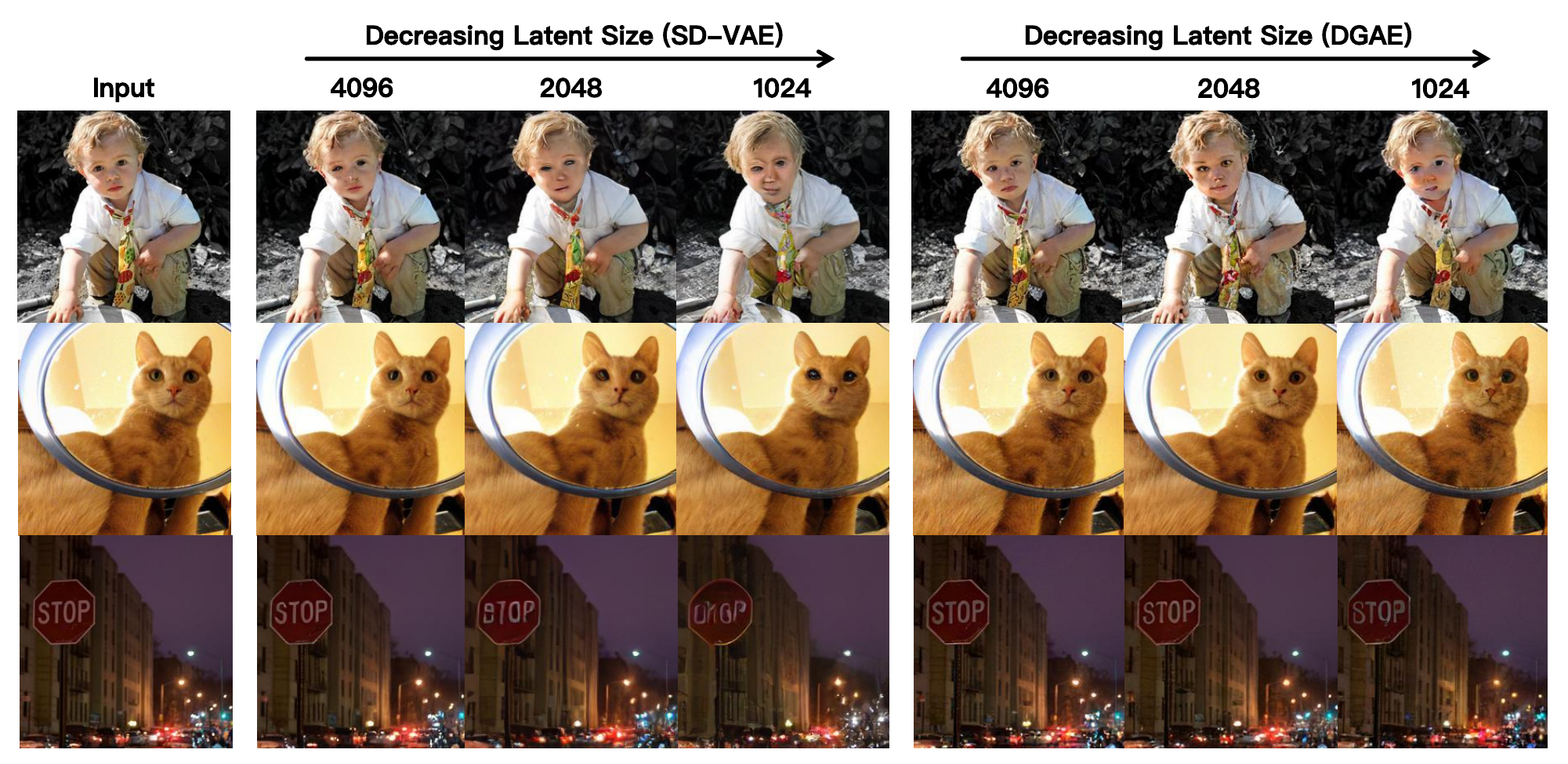}
  %\vspace{-13pt}
  \caption{\textbf{Reconstruction samples with different latent sizes.} The result was obtained under a fixed spatial compression rate of f16, with the channel dimension of the latent representation gradually decreased. As the latent size decreases, SD-VAE tends to collapse, while {\ours} still maintains a high fidelity.}
  
  \label{fig:latent_size}
  %\vspace{-10pt}
\end{figure*}

\begin{figure*}
  \centering
  \includegraphics[width=1.0\linewidth,height=0.7\textheight, keepaspectratio]{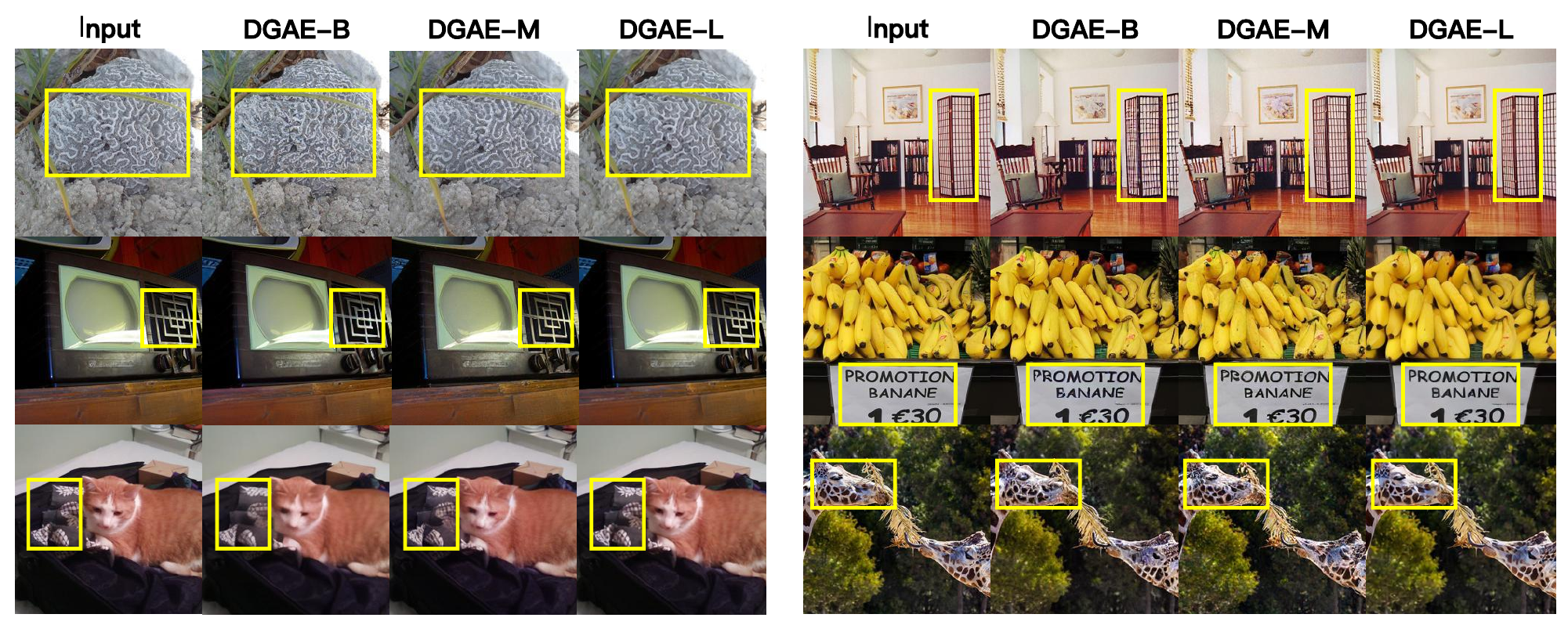}
  \caption{Scalability Evaluation of {\ours}. By scaling up the decoder, {\ours} achieves better reconstruction quality with enhanced detail preservation.}
  \label{fig:sample_dg_scale}
  \vspace{-2pt}
\end{figure*}
\begin{table*}
  \centering
  \caption{\textbf{Reconstruction Results of Scaled-Up {\ours} on ImageNet {\boldmath$256 \times 256$}}. A larger decoder architecture (Unet) in {\ours} leads to improved quantitative reconstruction results.}%
  \vspace{-10pt}
  \begin{tabular}{cccccccc}
    \toprule
     \multirow{2}{*}{\textbf{Autoencoder}} &\multirow{2}{*}{\textbf{Total Params}}& \multicolumn{3}{c}{\textbf{Decoder Arch}}  & \multicolumn{3}{c}{\textbf{Reconstruction Performance}}   \\
     \cmidrule(lr){3-5}
     \cmidrule(lr){6-8}
     &&\textbf{$\mathbf{Unet_{channel}}$}&\textbf{$\mathbf{t_{emb}}$}&\textbf{Params}&\textbf{rFID-5k$\downarrow$}& \textbf{PSNR$\uparrow$} & \textbf{SSIM$\uparrow$}\\
    \midrule
    {\ours}-B&157M&128&512&122M& 4.23& 25.73 & 0.76   \\
    {\ours}-M&310M&192&768&276M& 3.76& 26.10 & 0.76  \\
    {\ours}-L&525M&256&1024&491M& 3.58& 26.13 & 0.77  \\
    \bottomrule
  \end{tabular}
  \label{tab:rfid_dg_scale}
  \vspace{-10pt}
\end{table*}

\begin{table*}
  \centering
  \caption{{\ours} effectively improves the reconstruction performance. As the latent size decreases, the performance of SD-VAE drops significantly, while {\ours} remains relatively stable. In addition, during our reproduction of SD-VAE-f32, we found that the GAN was highly prone to collapse, and we did not obtain a reasonable result.}
  \vspace{-5pt}
  \begin{tabular}{@{}ccccccc@{}}
    \toprule
      \multirow{2}{*}{\textbf{Spatial downsampling}} & \multirow{2}{*}{\textbf{Latent Shape}} & \multirow{2}{*}{\textbf{Latent Size}} &\multirow{2}{*}{\textbf{Autoencoder}}& \multicolumn{3}{c}{\textbf{Reconstruction Performance} }  \\
      \cmidrule(lr){5-7}
      &&&&\textbf{rFID-5k$\downarrow$}& \textbf{PSNR$\uparrow$} & \textbf{SSIM$\uparrow$}  \\
    \midrule
     \multirow{2}{*}{f8} & \multirow{2}{*}{$ 32\times32\times4 $ }&\multirow{2}{*}{4096 } &SD-VAE & 4.91 &24.60& 0.74  \\
     &&&{\ours}& 4.23 &25.73& 0.76 \\
     
    \midrule
     \multirow{6}{*}{f16} &\multirow{2}{*}{$ 16\times16\times16 $}&\multirow{2}{*}{4096 } &SD-VAE& 4.62 &23.90&0.75 \\
     &&&{\ours} & 3.98 &25.33&0.76\\
     \cmidrule(){2-7}
     &\multirow{2}{*}{$ 16\times16\times8 $} &\multirow{2}{*}{2048 }&SD-VAE& 8.53 &22.62&0.70 \\
     &&& {\ours}& 4.99 &24.46&0.74\\
     \cmidrule(){2-7}
     &\multirow{2}{*}{$ 16\times16\times4 $} &\multirow{2}{*}{1024 }&SD-VAE& 16.02 &20.97&0.63 \\
      &&&  {\ours} & 9.45 &21.54&0.68\\
      
    \midrule
     \multirow{2}{*}{f32} &\multirow{2}{*}{$ 8\times8\times64 $}  &\multirow{2}{*}{4096 }&SD-VAE& 9.07 &22.26&0.69\\
     &&& {\ours} & 5.14 &23.07&0.73\\

    \bottomrule
  \end{tabular}
  
  \label{tab:rfid_spatial}
\end{table*}
\begin{table*}[t!]
  \centering
  \caption{Class-conditional generation results on ImageNet {\boldmath$256\times256$} (w/o CFG). With the latent representations learned by {\ours}, DiT achieves comparable generation quality using only half the latent dimensionality of SD-VAE. (In specification, 'f' and 'c' represent the spatial downsampling and channel, respectively.) }
  \vspace{-5pt}
  \begin{tabular}{cccccccc} 
    \toprule
     \multirow{2}{*}{\textbf{Diffusion Model}} & \multicolumn{3}{c}{\textbf{Autoencoder}} & 
     \multicolumn{4}{c}{\textbf{Generation Performance }} \\
     \cmidrule(lr){2-4}
     \cmidrule(lr){5-8}
     &\textbf{Latent Size}&\textbf{Specification}&\textbf{Type}&\textbf{gFID$\downarrow$} & \textbf{sFID$\downarrow$} & \textbf{Precision$\uparrow$}  & \textbf{Recall $\uparrow$}\\
     
    \midrule
    \multirow{4}{*}{DiT-XL {\cite{william2023dit}}}
    &  \multirow{2}{*}{4096}&\multirow{2}{*}{f16c16}& SD-VAE & 11.51 & 5.47 &0.67 & 0.65 \\
    &  && {\ours} & 10.41 & 5.39 & 0.71 & 0.61 \\
    \cmidrule(lr){2-8}
    &  \multirow{2}{*}{2048}&\multirow{2}{*}{f16c8}& SD-VAE & 12.49 &5.48 & 0.68 & 0.62\\
    &  && {\ours} & 11.16 & 5.43& 0.69 & 0.62\\
    \bottomrule
  \end{tabular}
  
  \label{tab:gfid_dg_vs_sd}
\end{table*}
%-------------------------------------------------------------------------\
\subsection{Setup}
\textbf{Implementation Details.} The encoder architecture of {\ours}  remains consistent with that of SD-VAE, while the decoder follows the standard convolutional U-Net architecture of ADM. We implemented three different latent sizes: 4096, 2048, and 1024, with each size corresponding to a distinct spatial compression ratio. For conditional denoising in the decoder, we first upsample the latent representation to pixel level using nearest-neighbor interpolation, and then concatenate it with random noise along the channel dimension.

\label{sec:setup}
\textbf{Data.} All reconstruction and generation experiments are conducted on the ImageNet-1K dataset~\cite{Jia2009ImageNet} to evaluate the performance of {\ours}. We preprocess all images by resizing them to a resolution of $256 \times 256$ pixels. During training, we apply standard data augmentation techniques, including random cropping and random horizontal flipping, to encourage robustness and improve generalization. For evaluation, center cropping is used to ensure stable and consistent results.

\textbf{Baseline.} To assess the effectiveness of our approach, we compare it with SD-VAE~\cite{rombach2022stable_diffusion}, a widely adopted baseline in visual generation. Compared to the single-step decoding of SD-VAE, {\ours} conditions on the latent representation and performs multi-step denoising of Gaussian noise to recover the original image. Aside from architectural differences, both models are trained under identical settings to ensure a fair and controlled comparison.

\textbf{Training.} We train all models with a batch size of $96$, matching the configuration used in SD-VAE. All model parameters are randomly initialized and optimized using AdamW ($\beta_1 = 0.9$, $\beta_2 = 0.95$, $\epsilon=1\mathrm{e}{-8}$) with an initial learning rate of $1\mathrm{e}{-4}$, linearly warmed up for the first 10K steps and decayed to $1\mathrm{e}{-5}$ using a cosine scheduler. We apply a weight decay of $0.1$ for regularization and clip gradients by a global norm of $1.0$. To accelerate training, we adopt mixed-precision training with bfloat16.

\textbf{Evaluation.} We employ a range of metrics to comprehensively evaluate both reconstruction and generation performance. For reconstruction, we report PSNR and SSIM~\cite{hore2010image} to assess pixel-wise accuracy and perceptual similarity, respectively. Additionally, we adopt the Fréchet Inception Distance (rFID)~\cite{Martin2017FID}, computed between the original and reconstructed images, as a more perceptually aligned metric. Notably, we use the rFID, calculated on a fixed subset of 5K images from the ImageNet-1K~\cite{Jia2009ImageNet} validation set. For generation, we evaluate the synthesized samples using several standard metrics~\cite{dhariwal2021diffusion}: generation FID (gFID), sFID~\cite{nash2021generating}, Precision, and Recall. These metrics collectively measure fidelity, diversity, and sample quality of the generated outputs, providing a thorough assessment of the generative capabilities of our model.

\subsection{The reconstruction capability of {\ours}}
\label{sec:rec}
We first demonstrate that {\ours} achieves better reconstruction results with higher spatial compression rates and smaller latent sizes, proving its ability to learn more expressive latent representations. Then, as discovered in SD-VAE, scaling up the decoder can effectively enhance the reconstruction performance of {\ours}. Unless otherwise specified, {\ours}-B is used by default in the experiments of this section.

\textbf{Spatial Compression.} To verify whether the Diffusion Model can mitigate the performance degradation under high spatial compression rates like GAN, we test {\ours} in latent spaces with various spatial compression rates. As shown in \cref{tab:rfid_spatial}, {\ours} achieves superior performance across all spatial compression rates. Qualitatively, as shown in \cref{fig:sample_dgvae_vs_sdvae}, we find that {\ours} is capable of modeling better texture features and symbols. The results further confirm that the encoder has already stored the semantics of the image in the latent representation, and what we need to do is to uncover it. 

\textbf{Latent Compression.} Moreover, under higher spatial compression rates, increasing the number of latent channels can improve the reconstruction performance of autoencoders. However, this comes at the cost of significantly larger diffusion models and more challenging optimization~\cite{yao2025reconstruction}. This motivates the use of a more compact latent space.
We investigate the ability of {\ours} to mine information by fixing the spatial compression rate and reducing the number of channels in the latent representation. As shown \cref{tab:rfid_spatial}, the gap between {\ours} and SD-VAE widens as the latent size decreases. In addition to the quantitative results, \cref{fig:latent_size} shows image reconstruction samples produced by SD-VAE and DC-AE. The reconstructed images by {\ours} demonstrate better visual quality than those reconstructed by SD-VAE. In particular, for autoencoders with a latent size of 1024, {\ours} still maintains good visual quality for small text and human faces.

\textbf{Scalability.} As emphasized in \cref{sec:intro}, the decoder plays a central role in autoencoder architectures. To evaluate the scalability of {\ours}, we fix the encoder and progressively scale the decoder. Specifically, we construct three variants with increasing model capacities: {\ours}-B, {\ours}-M, and {\ours}-L. The number of parameters and detailed configurations of the corresponding U-Net decoders are provided in \cref{tab:rfid_dg_scale}. As shown in \cref{fig:sample_dg_scale}, larger decoders significantly enhance the model’s ability to capture structural and fine-grained image details. Quantitative results in \cref{tab:rfid_dg_scale} further support this observation, demonstrating that {\ours} is a scalable and effective autoencoder framework.

While good latent representations should enable faithful reconstruction from the pixel space, its true utility lies in how effectively it supports downstream generative modeling. In particular, a meaningful and compact latent space should facilitate the training of powerful diffusion models. Therefore, in \cref{sec:gen}, we evaluate whether the representations learned by {\ours} contribute to improved image synthesis performance.

\subsection{Latent Diffusion Model}
\label{sec:gen}
We compare the performance of training a latent diffusion image generation model on two different latent representations, learned by {\ours} or SD-VAE. Specifically, we use DiT-XL/1~\cite{william2023dit} as the latent diffusion model for class-conditional image generation on ImageNet-1K\cite{Jia2009ImageNet}. In this section, our focus is to demonstrate the effectiveness of the latent representation learned by {\ours}. Therefore, we train the diffusion model for only 1M steps instead of the original 7M steps~\cite{william2023dit}.

\begin{figure}
  \centering
  \includegraphics[width=1.0\linewidth,height=0.7\textheight, keepaspectratio]{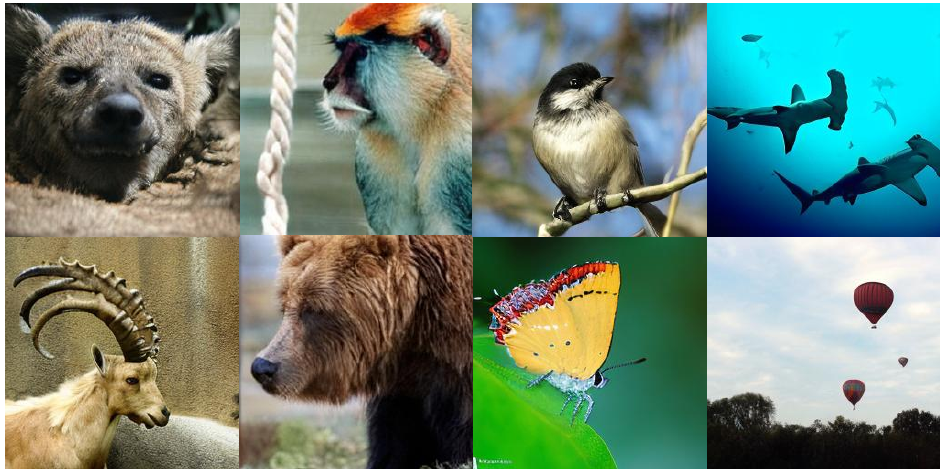}
  \caption{Class-Conditional Image Generation Results of DiT-XL Trained on ImageNet 256×256. Despite being trained for only 1M steps, DiT-XL achieves high-quality  generation on the {\ours}'s latent space.}
  \label{fig:sample_gen}
\end{figure}

As shown in \cref{tab:gfid_dg_vs_sd}, {\ours} consistently outperforms SD-VAE across different latent sizes. In particular, even with a latent dimensionality reduced by half, {\ours} still achieves superior generation quality, demonstrating the robustness of its latent space. \Cref{fig:sample_gen} shows samples generated by DiT trained on {\ours}'s latent representations with a size of 2048. Even after just 1M training steps, the model is able to produce visually compelling results.

\begin{figure}
  \centering
  \includegraphics[width=0.99\linewidth,height=0.7\textheight, keepaspectratio]{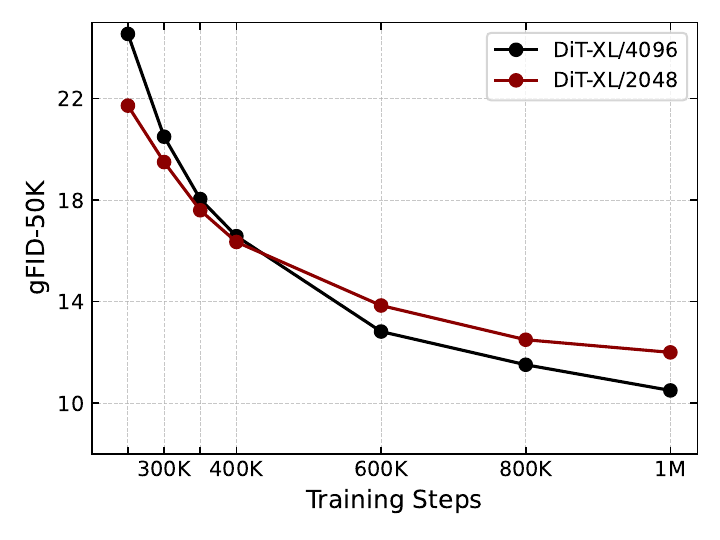}
  \caption{Convergence Curves of DiT-XL under Different Latent Sizes. As the training steps increases, the DiT-XL trained with a latent size of 2048 converges more quickly. }
  \label{fig:gfid}
  \vspace{-9pt}
\end{figure}

To further understand the benefits of smaller latent sizes, we examined the convergence behavior of DiT models with different latent sizes. As shown in \cref{fig:gfid}, DiT models converge more quickly with smaller latent sizes, indicating that diffusion models achieve faster convergence and reduced training costs when utilizing smaller latent dimensions.

Next, we explore why the latent representation learned by {\ours} is more effective.

\subsection{Latent Reprensentation}
\label{sec:latent_vis}
\begin{figure}[t!]
  \centering
  \includegraphics[width=1.0\linewidth,height=0.7\textheight, keepaspectratio]{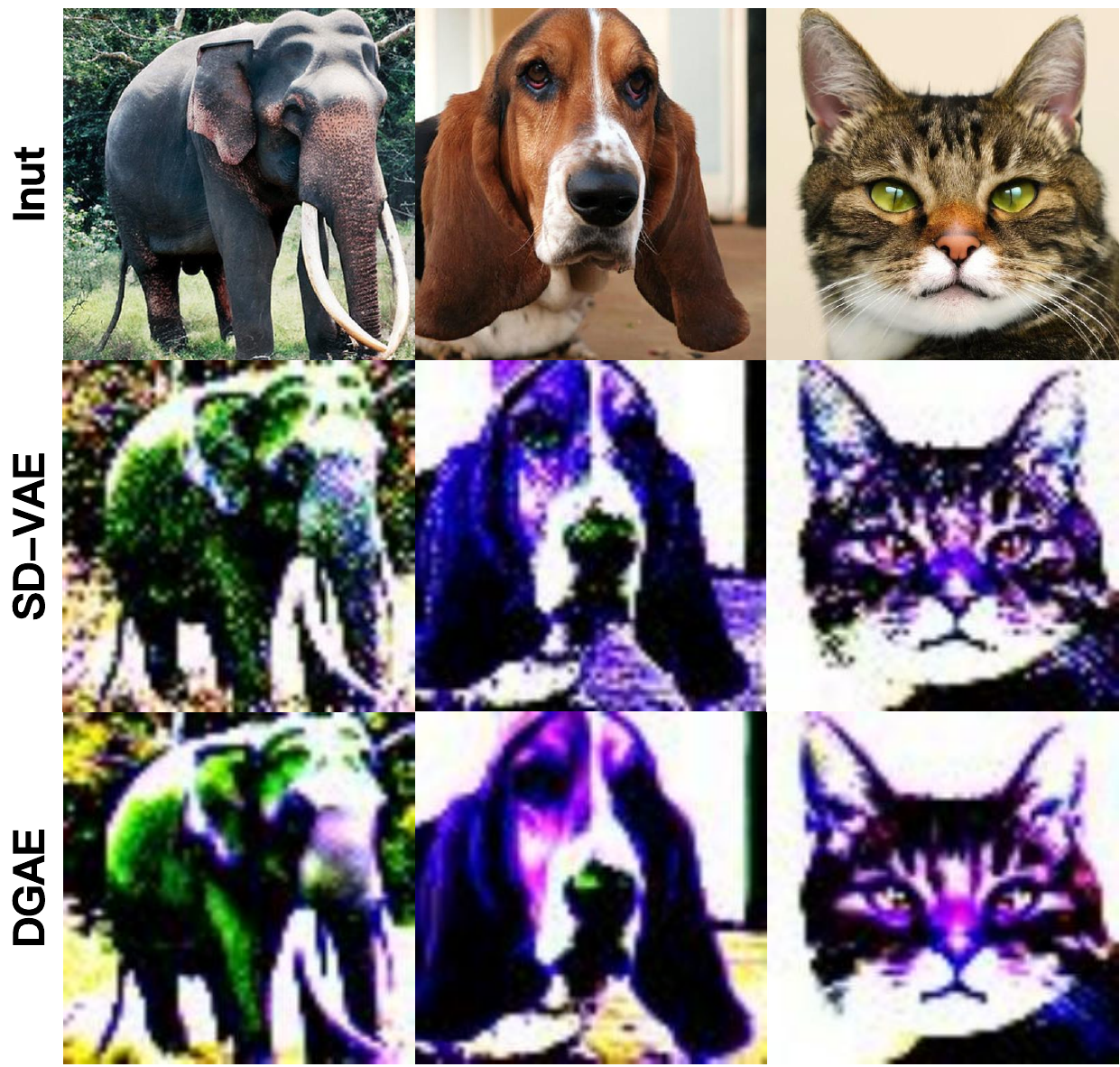}
  \caption{Visualizing the Latent Representations of SD-VAE and {\ours}. By applying a simple linear projection to map the latent representations to the RGB space, we observe that {\ours} exhibits a smoother latent space compared to SD-VAE, without compromising reconstruction performance.}
  \label{fig:latent_visual}
  \vspace{-9pt}
\end{figure}

The use of KL normalization in SD-VAE involves a trade-off between information capacity and detailed information~\cite{Tschannen2023GIVTGI}, this requires the decoder to be able to fill in the lost details. Diffusion models, however, possess a unique coarse-to-fine nature: they first synthesize low-frequency signal components and later refine them with high-frequency details. This property appears particularly well-suited for compensating the loss of fine-grained information in the latent space. 
As shown in \cref{fig:latent_visual}, we visualized the latent representations of {\ours} and SD-VAE separately and found an interesting phenomenon: {\ours} has a smoother latent space. This eliminates the burden of learning nonlinear relationships in the latent space for generative models. Based on the smoother latent representation, the decoder is freed up to fill in the details. This may be the reason why {\ours} can achieve better reconstruction results with a smaller latent space. 


\section{Related Work}
\label{sec:relatedwork}

\textbf{Diffusion models. }Diffusion models ~\cite{song2020sde, Arash2021sbinlatent, ho2020ddpm, Alexander2021iddpm, Tero2022edm, william2023dit, ma2024sit, kingma2021variational} have supplanted traditional generative models, including GANs ~\cite{Ian2014gan, Tero2021stylegan, he2019attgan, he2021eigengan} and VAEs ~\cite{kingma2014vae,  dai2019diagnosing, higgins2017beta, rezende2018taming}, emerging as the predominant framework in the realm of visual generation.  Due to direct optimization in the pixel space and the use of multi-timestep training and inference, diffusion models were initially applied only to the synthesis of low-resolution visual content~\cite{Arash2021sbinlatent,ho2020ddpm,Alexander2021iddpm,Tero2022edm}. To scale them up to high-resolution image generation, subsequent works either adopt super-resolution techniques to increase the generated images to higher resolutions ~\cite{ramesh2022dalle2, saharia2022imagen, ho2022cascaded} or perform optimization in the latent space instead of the pixel space~\cite{rombach2022stable_diffusion}. In parallel, another line of research focuses on accelerating the sampling process through methods such as knowledge distillation ~\cite{Tim2022ProgressiveDistillation, sauer2023sdxl_turbo, song2023consistency} and noise scheduling ~\cite{kingma2021variational, Alexander2021iddpm, kong2021fast}. By adopting these strategies, diffusion models have achieved remarkable results in visual generation~\cite{ho2022imagen_video, yang2024cogvideox, podell2024sdxl, blattmann2023svd_stablevideodiffusion, polyak2024moviegen, nvidia2025cosmos, saharia2022imagen, dong2023dreamllm, ma2025step, pku_opensora_plan, opensora, chen2024omnicreator, kahatapitiya2024adaptive, chen2024pixart_alpha, chen2024pixart_delta, chen2024pixart_sigma, xiao2024omnigen, xie2024sana, peng2025open, ma2024latte, chen2024videocrafter2, bar2024lumiere, bao2024vidu, qin2024xgen, xing2025dynamicrafter, zhang2023adding, xu2024ctrlora, liu2025step1x,zhou2025taming,xie2024gladcoder,tan2024waterdiff,peng2025dreambench}. 

\textbf{Visual Autoencoders. }Due to the success of LDM’s SD-VAE, substantial efforts have been devoted to developing better autoencoders. To enable a more efficient denoising process, follow-up works have focused on improving reconstruction accuracy under high spatial compression~\cite{chen2024dcae, esser2024sd3, dai2023emu, tian2024reducio, hacohen2024ltx}.

 Another popular trend is employ wavelet transforms to enhance high-frequency details ~\cite{pku_opensora_plan, nvidia2025cosmos, yu2025frequency}.In addition to the continuous AEs explored in this work, multiple discrete AEs~\cite{van2017neural, razavi2019generating, wang2024omnitokenizer, nvidia2025cosmos, tang2024vidtok, yu2024magvitv2, li2024imagefolder, zhao2024image, tian2024visual} are proposed to aid autoregressive tasks~\cite{sun2024autoregressive, han2024infinity, esser2021vqgan}.

However, the above methods all incorporate GAN loss as part of their training objective. While this enhances the autoencoder's ability to capture texture and structural details, it also introduces training instability. Moreover, these approaches primarily focus on improving reconstruction quality, while overlooking the importance of the latent size. To address these limitations, we propose using diffusion models, which provide stable training and better utilize compact latent spaces.

\textbf{Diffusion Autoencoders.} Early works incorporate diffusion decoders into autoencoders~\cite{shi2022divae, preechakul2022diffusion_ae, bachmann2025flextok} primarily aimed to leverage the stochastic nature of diffusion processes to enhance image quality, without establishing a clear connection to the latent diffusion modeling (LDM) framework~\cite{rombach2022stable_diffusion}. SWYCC~\cite{birodkar2024swycc} refines the output of an autoencoder by appending a post-hoc diffusion module, while $\epsilon$-VAE~\cite{zhao2024eps_vae} integrates diffusion decoders directly within the LDM paradigm. Parallel to our work, DiTo~\cite{chen2025dito} introduces a diffusion-based autoencoder for self-supervised learning, with a stronger focus on architectural scalability.

\section{Conclusion}
\label{sec:conclusion}
We demonstrate that the decoder plays a more important role than the encoder in autoencoders. By introducing a diffusion process to assist the decoder in image reconstruction, our {\ours} can map images to a smaller latent size without a decrease in precision. Moreover, compared to SD-VAE that employs a GAN, the training process of {\ours} is more stable. In addition, we find that diffusion models can converge more quickly at a smaller latent size.

%\newpage
\bibliographystyle{ACM-Reference-Format}
\bibliography{ref}

\end{document}